# Enhanced Techniques for PDF Image Segmentation and Text Extraction

**D.Sasirekha**
Research Scholar, Computer Science
Karpagam University
Coimbatore, Tamilnadu, India
dsasirekha@gmail.com

**Dr.E.Chandra,**
Director, Dept of Computer Science
Dr.SNS Rajalakshmi college of Arts and Science,
Coimbatore,Tamilnadu,India
crcspeech@gmail.com

*Abstract*— Extracting text objects from the PDF images is a challenging problem. The text data present in the PDF images contain certain useful information for automatic annotation, indexing etc. However variations of the text due to differences in text style, font, size, orientation, alignment as well as complex structure make the problem of automatic text extraction extremely difficult and challenging job. This paper presents two techniques under block-based classification. After a brief introduction of the classification methods, two methods were enhanced and results were evaluated. The performance metrics for segmentation and time consumption are tested for both the models.

## I. INTRODUCTION

With the drastic advancement in Computer Technology & communication technology, the modern society is entering to the information edge. In change in the traditional document system (paper etc), people now follow electronic document system (PDF Format) for communication and storage which is currently imperative. But on complex matters, the document image is difficult to accurately identify the information directly out of the need. On such cases preprocessing the document is done before its entry. Image segmentation theory, as digital image processing has become an important part of people active research.

Image processing document image segmentation theory is an important research topic in the process it is mainly between the document image pre-processing and advanced character recognition an important link between. The relatively effective and commonly used for document image segmentation and classification methods include threshold, and geometric analysis and other categories.

After segmenting, Text part is detected and extracted for further process, earlier, text extraction techniques have been developed only on monochrome documents [1]. These techniques can be classified as bottom-up, top-down and hybrid. Later with the increasing need for color documents, techniques [2]have been proposed

The segmentation techniques like Block based image segmentation [3] is used extensively in practice. Under this block based segmentation, the comparison goes (i) AC-Coefficient Based technique and (ii) Histogram Based technique

This paper is organized as follows: In section II the brief introduction of PDF Image. Section III discuss about the review of block based segmentation. Section IV discusses in detail about the Text extraction using proposed techniques. Section V discuss about the experimental results of the two models. Finally the section 6 concludes the paper.

## II. PDF IMAGE

PDF format is converted into images using available commercial software's so that each PDF page is converted into image format. From that image format the text part are segmented and extracted for further process.

## III. BLOCK BASED SEGMENTATION

The goal of segmentation is to simplify and/or change the representation of an image into something that is more meaningful and easier to analyze. Image segmentation is typically used to locate objects and boundaries (lines, curves, etc.) in images. More precisely, image segmentation is the process of assigning a label to every pixel in an image such that pixels with the same label share certain visual characteristics.

Most of the recent researches in this field mainly based on either layer based or block based. This block based segmentation approach divides an image into blocks of regions (Fig:1). Each region follows approximate object boundaries, and is made of rectangular blocks. The size of the blocks may vary within the same region to better approximate the actual object boundary.

Block-based segmentation algorithms are developed mostly for grayscale or color compound image, For example ,In [4], text and line graphics are extracted from check images. In [5],proposed block based clustering algorithm, [6] propose a classification algorithm, based on the threshold of the number of colors in each block. [7], approach based on available local texture features[8], detection using mask [9], block classification algorithm for efficient coding by thresholding DCT energy [10] with better segmentation and were used during further experimentation.

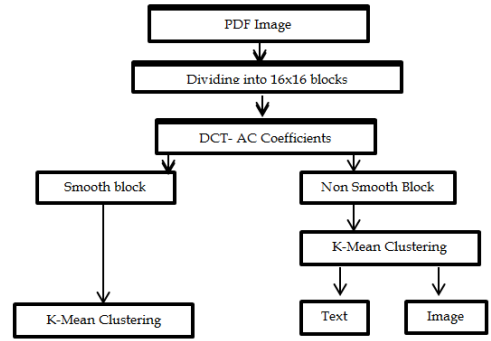

Figure 2: AC Coefficient based segmentation

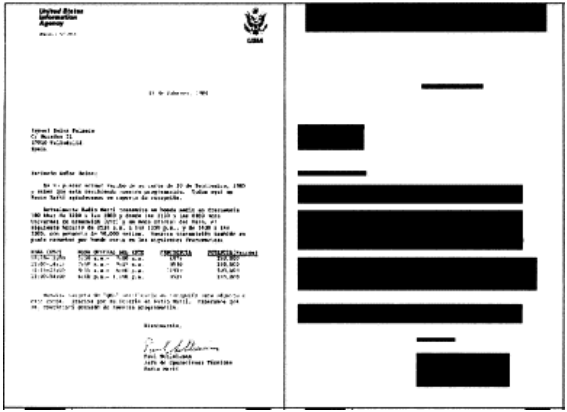

Figure 1: Block Based Segmentation

The following sub sections discuss the two techniques (A) AC Coefficient based technique & (B) Histogram based technique in block based segmentation

*A. AC Coefficient based technique*

The first model uses the AC coefficients introduced during (Discrete Cosine Transform)DCT to segment the image into threeblocks, background, Text and image blocks[11] [12] [14][15]. The background block has smooth regions of the imagewhile the text / graphics block has high density of sharp edges regions and image block has the non-smooth partof the PDF image( Fig:2). AC energy is calculated from AC coefficients and is combined with a user-definedthreshold value to identify the background block initially. The AC energy of a block 's' is calculated using Equation (1)

$$E_s = \sum_{i=1}^{63} Y_{s,i}^2 \quad (1)$$

where $Y_{s,i}$ is the estimate of the i-th DCT coefficient of the block 's', produced by JPEG decompression. When the $E_s$ value thus calculated is lesser than a threshold $T_1$, then it is grouped as smooth region; else it is grouped as non-smooth region. After much experimentation with different images, the thresholds $T_1$ and $T_2$ with 20 and 70 respectively resulted

To further identify the image and text regions of the compound image, the non-smooth blocks are considered and a 2-D feature vector is computed from the luminance channel. Two feature vectors $D_1$ and $D_2$ are determined.

It was reported by Konstantinides and Tretter (2000)[17] that the code lengths of text blocks after entropy encoding tend to be longer than non-text blocks due to the higher level of high frequency content in these blocks. Thus, the first feature, $D_1$, calculates the encoding length using Equation (3).

$$D_{s,1} = \frac{1}{64}\left(f(Y_{s,0} - Y_{s-1,0}) + \sum_{1}^{63} f(Y_{s,i})\right) \quad (2)$$

Where $f(x) = \begin{cases} \log_2(|x|) + 4 & \text{if } |x| > 1 \\ 0 & \text{Otherwise} \end{cases}$

The second feature, $D_2$, is the measure of how close a block is to a two-colored block. For each block 's', a two-color projection is performed on the luminance channel. Each block is clustered in to two groups using k-means clustering algorithms with means denoted by $\theta_{s,1}$ and $\theta_{s,2}$. The two-color projection is formed by clipping each luminance of each pixel to the mean of the cluster to which the luminance value belongs. The $l^2$ distance between the luminance of the block and its two-color projection is then a measure of how closely the block resembles a two-color block. This projection error is normalized by the square of the difference of the two estimated means, $|\theta_{s,1} - \theta_{s,2}|^2$, so that a high contrast block has a higher chance to be classified as a text block. The second feature is then calculated as

$$D_{s,2} = \frac{1}{|\theta_{s,1} - \theta_{s,2}|^2} \sum_{i=0}^{63} |X_{s,i} - X'_{s,i}|^2 \quad (3)$$

where $X_{s,i}$ is the estimate of the $i^{th}$ pixel of the block 's', produced by JPEG decompression and $X'_{s,i}$ is the value of the $i^{th}$ pixel of the two-color projection. If $\theta_{s,1} = \theta_{s,2}$, then $D_{s,2} = 0$.

For the norm used in k-means clustering, the contributions of the two features are differently weighted. Specifically, for a vector $D^t = [D1,D2]$, the norm is calculated by

$$\| D \| = \sqrt{D_1^2 + \gamma D_2^2} \quad (4)$$

where $\gamma = 15$. All clusters whose mean value is greater than $D_{s,1}$ and lesser than $D_{s,2}$ are grouped as text blocks. The rest of the blocks are termed as picture blocks.

Each cluster is fit with a Gaussian mixture model. The two models are then employed within the proposed segmentation algorithm (Bouman and Shapiro, 1994)[18] in order to classify each non-background image block as either a text block or a image block. The SMAP algorithm is shown in Fig: 3.

---

1. Set the initial parameter values for all n', $\theta_{n,0} = 1$ and $\theta_{L-1,1} = 0.5$.
2. Compute the likelihood functions and the parameters $\theta_{n,0}$.
3. Compute $x^{(L)}$ using Equation

   $$x_s^{(L)} = \arg \max_{1 \le k \le M} l_s^{(L)}(k)$$

   and

   $$x_s^{(L)} = \arg \max_{1 \le k \le M} \left\{ l_s^{(L)}(k) + \log p_{x^{(n)}|x_{\partial s}^{(n+1)}}(k|x_{\partial s}^{(n+1)}) \right\}$$

4. For scales n = L-1 to n = 0
   a) Use EM algorithm to iteratively compute $\theta_{n,1}$ and T. Subsample by $P^{(n)}$ when computing T and stop when
   b) compute $\theta_{n,o}$ using Equation .

   $$\theta_{n,0} = \frac{\sum_{h=0}^{2} T_{i,h}}{\sum_{i=0}^{1} \sum_{h=0}^{2} T_{t,h}}$$

   c) Compute $x^{(n)}$
   d) Set $\theta_{n-1,1} = \theta_{n,1(1-log_6 2)}$
5. Repeat steps 2 and through 4.

---

Figure 3: SMAP Algorithm

*B. Histogram Based Technique*

The second block-based segmentation model sequences a histogram-based threshold approach [13][16]. In this technique the image is segmented using a series of rules. The segmentation process involves a series of decision rules from the block type with the highest priority to the block type with the lowest priority. The decision for smooth and text blocks is relatively straightforward. The histogram of smooth or text blocks is typically dominated by one or two intensity values (modes). Separating the Text and image blocks from the PDF image is challenging.

Here the intensity value is defined as mode if its frequency satisfies two conditions,

(i) it is a local maximum and
(ii) the cumulative probability around it is above a pre-selected threshold, T.

The algorithm begins by calculating the probability of intensity value i, where i = 0... 255 using Equation (2)

(iii) $\quad pi = freq(i) / B^2 \quad (5)$
(iv)

Where B is the block size and a value of 16 is used in the experiment. Then the mode $(m_1, \ldots, m_x)$ is calculated and the cumulative probability around the mode m is computed using Equation (6).

$$c_n = \sum_{m-A}^{m+A} p_i \quad -(6)$$

The decision rules used is given in Figure.4.

---

Rule 1 : If N = 1 and $c_1 > T_1 \rightarrow$background block
Rule 2 : If N = 2 and $c_1 + c_2 > T_1$ and $|C_1-C_2| > T_2 \rightarrow$Text block
Rule 3 : If N ≤ 4 and $c_1+c_2+c_3+c_4 > T_1 \rightarrow$Graphics block
Rule 4 : If N > 4 and $c_1+c_2+c_3+c_4 < T_3 \rightarrow$Picture block

---

Figure 4: Decision Rules

In the above said Decision rules, after many tests, the thresholds $T_1$, $T_2$ and $T_3$ were set as 30, 45 and 70 respectively, getting better results.

The following section 4 deals with the text extraction techniques.

IV. TEXT EXTRACTION

By applying the two techniques of block based method, the image is segmented into
1. Smooth region (Background)
2. Non Smooth region
   I. Text regions
   II. Image region

The technique (2.1) indicate that while segmenting the PDF image, background is identified as smooth blocks. The foreground (non smooth block), using K-means algorithm the text and image blocks are segmented and thus text part is separated or extracted from the PDF image.

In the technique (2.2), the PDF image is segmented into 16 X 16 blocks, then a histogram distribution for each pixel in each segmented group is computed. Grouping of pixels is done low, mid and high gradient pixels. Threshold value is assigned to calculate the value to identify the text block and image block (4)

if High gradient pixels + low gradient pixels < T1 then
    *the block is image*
else if (High gradient pixels < T2 && low gradient pixels >T3)then
if number of colour level < T4 then
    *the block is Text block*
**else**
*the block is image*
**End**
*(Where T1=50; T2=45; T3=10; T4=4)*

## V. EXPERIMENTAL RESULTS

The following fig: 5 is the combination of PDF images, which are used for testing.

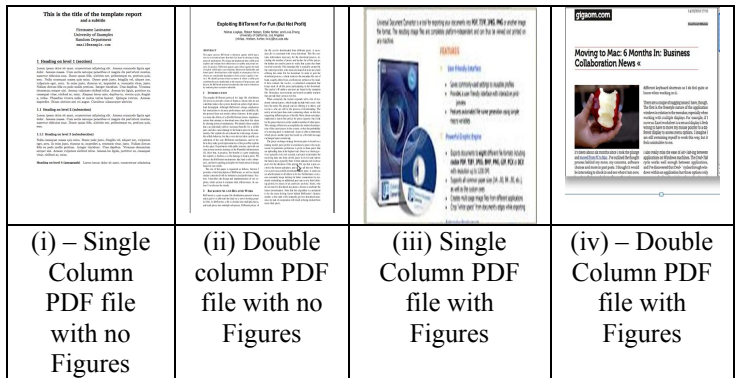

| (i) – Single Column PDF file with no Figures | (ii) Double column PDF file with no Figures | (iii) Single Column PDF file with Figures | (iv) – Double Column PDF file with Figures |

Fig 5: sample PDF files used for testing

**Total No. of PDF Files used for testing – 100 PDF files**

    20 – Single Column Files with no Figures
    20 – Double column Files with no Figures
    30 – Single Column Files with Figures
    30 – Double Column Files with Figures
    Evaluation Method: 10-fold cross validation technique

The Table 1 shows the comparison rate of the two proposed methods.

| Metrices | Single column PDF image with no figures | | Double column PDF image with no figures | | Single column PDF image with figures | | Double column PDF image with figures | |
|---|---|---|---|---|---|---|---|---|
| | AC coefficient based | Histogram based | AC coefficient based | Histogram based | AC coefficient based | Histogram based | AC coefficient based | Histogram based |
| Accuracy | 94.33 | 93.87 | 92.66 | 91.87 | 93.51 | 92.44 | 90.19 | 91.67 |
| False positive | 5.67 | 6.13 | 7.34 | 8.13 | 6.49 | 7.56 | 9.81 | 8.33 |
| Time (seconds) | 20.71 | 14.91 | 22.57 | 13.57 | 26.64 | 13.06 | 21.02 | 13.10 |

Table1: comparison rate of the two proposed methods.

From the above table it shows that the accuracy rate is better in AC-coefficient based technique where the time consumption is more in this technique. Whereas the time consumption is less and the accuracy rate is better in Histogram based technique.

## VI. CONCLUSION

On seeing the advantage and disadvantage of both the algorithms, From the performance analysis, If the user is willing to trade a little time for a better accuracy, then the AC-Coefficient based technique will be suitable. However if the user requires quick retrieval and is willing to tolerate a slightly less reliable outcome, then the Histogram based technique is more suitable.

AUTHORS PROFILE

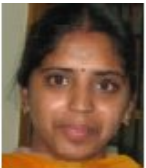

D.Sasirekha , completed her BSc (CS)-2003 in Avinashlingam University for Women, coimbatore and M.Sc (CS)-2005 in Annamalai University, Currently doing Ph.D (PT) (CS) in Karpagam University, Coimbatore and working in Avinashilingam University for Women, Coimbatore ,India.

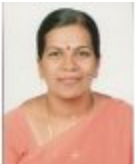

Dr.E.Chandra received her B.Sc., from Bharathiar University, Coimbatore in 1992 and received M.Sc., from Avinashilingam University ,Coimbatore in 1994. She obtained her M.Phil., in the area of Neural Networks from Bharathiar University, in 1999. She obtained her PhD degree in the area of Speech recognition system from Alagappa University Karikudi in 2007. She has totally 16 yrs of experience in teaching including 6 months in the industry. At present she is working as Director, School of Computer Studies in Dr.SNS Rajalakshmi College of Arts & Science, Coimbatore. She has published more than 30 research papers in National, International journals and conferences in India and abroad. She has guided more than 20 M.Phil., Research Scholars. At present 3 M.Phil Scholars and 8 Ph.D Scholars are working under her guidance. She has delivered lectures to various Colleges in Tamil Nadu & Kerala. She is a Board of studies member at various colleges. Her research interest lies in the area of Neural networks, speech recognition systems, fuzzy logic and Machine Learning Techniques. She is a Life member of CSI, Society of Statistics and Computer Applications. Currently Management Committee member of CSI Coimbatore chapter.